%% file: main.tex
\documentclass[letterpaper]{article} 
\usepackage{aaai23}
\usepackage{cite}
\usepackage{amsmath,amssymb,amsfonts}
\usepackage{graphicx}
\usepackage{balance}
\usepackage{textcomp}
\usepackage{xcolor}
\usepackage{multirow}
\usepackage{numprint}
\usepackage{pgfplots}
\usepackage{siunitx}
\usepackage{times}  
\usepackage{helvet}  
\usepackage{courier}  
\usepackage[hyphens]{url}  
\usepackage{graphicx} 
\def\UrlFont{\rm}  
\usepackage{natbib}  
\usepackage{caption} 
\usepackage{algorithm}
\usepackage[noend]{algpseudocode}

\pgfplotsset{every axis/.append style={
                    label style={font=\huge},
                    tick label style={font=\huge},
                    legend style={font=\huge}  
                    }}

\usepackage{newfloat}
\usepackage{listings}
\DeclareCaptionStyle{ruled}{labelfont=normalfont,labelsep=colon,strut=off} 
\floatstyle{ruled}
\newfloat{listing}{tb}{lst}{}
\floatname{listing}{Listing}
\title{
Complement Sparsification: Low-Overhead Model Pruning \\for Federated Learning
}
\author {
    Xiaopeng Jiang,
    Cristian Borcea
}
\affiliations {
Department of Computer Science, New Jersey Institute of Technology, Newark, NJ, USA\\
    xj8@njit.edu, borcea@njit.edu
}

\begin{document}
\maketitle

\begin{abstract}

Federated Learning (FL) is a privacy-preserving distributed deep learning paradigm that involves substantial communication and computation effort, which is a problem for resource-constrained mobile and IoT devices. Model pruning/sparsification develops sparse models that could solve this problem, but existing sparsification solutions cannot satisfy at the same time the requirements for low bidirectional communication overhead between the server and the clients, low computation overhead at the clients, and good model accuracy, under the FL assumption that the server does not have access to raw data to fine-tune the pruned models. We propose Complement Sparsification (CS), a pruning mechanism that satisfies all these requirements through a complementary and collaborative pruning done at the server and the clients.  At each round, CS creates a global sparse model that contains the weights that capture the general data distribution of all clients, while the clients create local sparse models with the weights pruned from the global model to capture the local trends. For improved model performance, these two types of complementary sparse models are aggregated into a dense model in each round, which is subsequently pruned in an iterative process. CS requires little computation overhead on the top of vanilla FL for both the server and the clients. We demonstrate that CS is an approximation of vanilla FL and, thus, its models perform well. We evaluate CS experimentally with two popular FL benchmark datasets. CS achieves substantial reduction in bidirectional communication, while achieving performance comparable with vanilla FL. In addition, CS outperforms baseline pruning mechanisms for FL. 

\end{abstract}

\input{introduction}
\input{related}
\input{methodology}
\input{results}

\input{conclusion}


\bibliography{sample-base}

\end{document}

%% file: introduction.tex
\section{Introduction}
Federated Learning (FL) is a collaborative deep learning paradigm that preserves user privacy. The clients send only locally-trained gradients to an aggregation server, without sharing their privacy-sensitive raw data. Traditionally, FL uses dense and over-parameterized deep learning (DL) models. Empirical evidence suggests that such models are easier to train with stochastic gradient descent (SGD) than more compact representations~\cite{kaplan2020scaling}.  
However, the over-parameterization comes at the cost of significant memory, computation, and communication overhead. This is a problem for resource-constrained mobile and IoT devices~\cite{lane2015early}, a major target for FL, which need to perform not only inference but also training. Therefore, reducing the computation and communication overhead in FL, while maintaining good model performance, is essential to ensure widespread FL deployment on mobile and IoT devices.
 
One potential solution to this problem is model pruning/sparsification, which aims to produce sparse neural networks without sacrificing model performance~\cite{10.5555/89851.89864}.
Sparse models result in significantly reduced memory and computation costs compared to their dense counterparts, while performing better than small dense models of the same size~\cite{han2015deep}. Sparse models lead to a better generalization of the networks~\cite{bartoldson2020generalization} and are more robust against adversarial attacks~\cite{10.1145/3386263.3407651,wu2020mitigating}. Pruning/sparsification can be used in FL, where the server and the clients can collaboratively optimize sparse neural networks to reduce the computation and communication overhead of training. 

Despite the benefits of sparse models, it is challenging to design a communication-computation efficient model pruning for FL. A typical pruning mechanism has three stages: training (a dense model), removing weights, and fine-tuning~\cite{https://doi.org/10.48550/arxiv.1810.05270}. Since a model with some of the weights removed has to recover the performance loss through additional fine-tuning in the back-propagation, the fine-tuning together with weights removal represents the computation overhead of the mechanism. In FL, this overhead cannot be placed only on the server because the server does not have access to the raw training data for fine-tuning. Therefore, pruning has to be done collaboratively between the server and the clients, and a significant computation overhead will be placed on the clients. 
Since FL exchanges model updates between the clients and the server every training round, smaller pruned models will lead to lower communication overhead. However, low communication overhead comes at the expense of computation overhead for pruning. 

To design a communication-computation efficient model pruning mechanism for FL, four requirements must be satisfied: (R1) reduce the size of the local updates from the clients to the server; (R2) reduce the size of the global model transferred from the server to the clients; (R3) reduce the pruning computation overhead at the clients; (R4) achieve comparable model performance with dense models in vanilla FL. All these requirements must be satisfied under the assumption that the server does not have access to raw data due to privacy concerns. 
None of the existing works on FL pruning~\cite{yu2021adaptive,jiang2022model,xu2021accelerating,horvath2021fjord,wen2022federated} can satisfy these requirements simultaneously. They either impose substantial computation overhead on the clients or only reduce the communication overhead from the clients to the server, but not vice versa.  The main unsolved problem is the apparent contradictory nature of the requirements.

This paper proposes {\bf Complement Sparsification (CS)}, a pruning mechanism for FL that fulfills all the requirements. The main idea is that the server and the clients generate and exchange sparse models complementarily, without any additional fine-tuning effort. 
The initial round starts from vanilla FL, where the clients train a dense model for the server to aggregate. The server prunes the aggregated model by removing low magnitude weights and transfers the global sparse model to the clients. 
In the following rounds, each client trains from the sparse model received from the server, and only sends back its locally computed sparse model. The client sparse model contains only the weights that were originally zero in the global sparse model, thus complementing the global model. Then, the server produces a new dense model by aggregating the client sparse models with the global sparse model from the previous round. As in the initial round, the server removes the weights with low magnitude and transfers the new global sparse model to the clients. The new model has a different subset of non-zero weights because the client model weights are amplified with a given aggregation ratio to outgrow other weights. In this way, all the weights in the model get updated to learn over time.

In CS, both the server and the clients transfer sparse models to save communication overhead bidirectionally (R1 and R2). Without deliberate fine-tuning, the computation overhead imposed on the system is minimized (R3). In CS, the pruning at the server preserves a global model that captures the overall data distribution, while the newly learnt client data distribution resides on the complementary weights (i.e., the zero weights of the global sparse model). 
Practically, the clients' training recovers the model performance loss during pruning without additional fine-tuning. Iteratively, the performance of the global model improves over time. Eventually, the clients can use the converged global sparse model for inference. 
This process can achieve comparable model performance with dense models in vanilla FL (R4).

We demonstrate that CS is an approximation of vanilla FL, and evaluate CS with two popular benchmark datasets~\cite{https://doi.org/10.48550/arxiv.1812.01097} for Twitter sentiment analysis and image classification (FEMNIST).
We measure model sparsity to quantify communication overhead. Specifically, CS achieves good model accuracy with server model sparsity between 50\% and 80\%. This sparsity represents the overhead reduction in the server-to-clients communication. The clients produce model updates with sparsity between 81.2\% and 93.2\%. The client sparsity represents the overhead reduction for client-to-server communication. CS reduces the computation overhead by 29.1\% to 49.3\% floating-point operations (FLOPs), compared with vanilla FL. We also demonstrate through experiments and a qualitative analysis that CS performs better than baseline model pruning mechanisms in FL~\cite{jiang2022model,xu2021accelerating} in terms of model accuracy and overhead.

%% file: related.tex
\section{Related Work}
\label{sec:related} 

Model pruning can be categorized as structured pruning and unstructured pruning.
There is a large body of literature on model pruning designed for centralized learning~\cite{guo2020multi,sanh2020movement,wang2020pruning}. There methods are computationally demanding and require a dataset representing the global data distribution. Therefore, they are not practical in FL, which does not share raw data with the server, and are difficult to use on resource-constrained mobile and IoT devices. Our CS model pruning, on the other hand, is designed for FL on resource-constrained devices. It does not requires a centralized dataset and eliminates explicit fine-tuning for computation efficiency. CS applies unstructured pruning in FL, due to its freedom to update different significant weights over FL training rounds and, thus, achieves better performance.

\begin{figure*}[t!]
  \centering
  \includegraphics[width=0.7\linewidth]{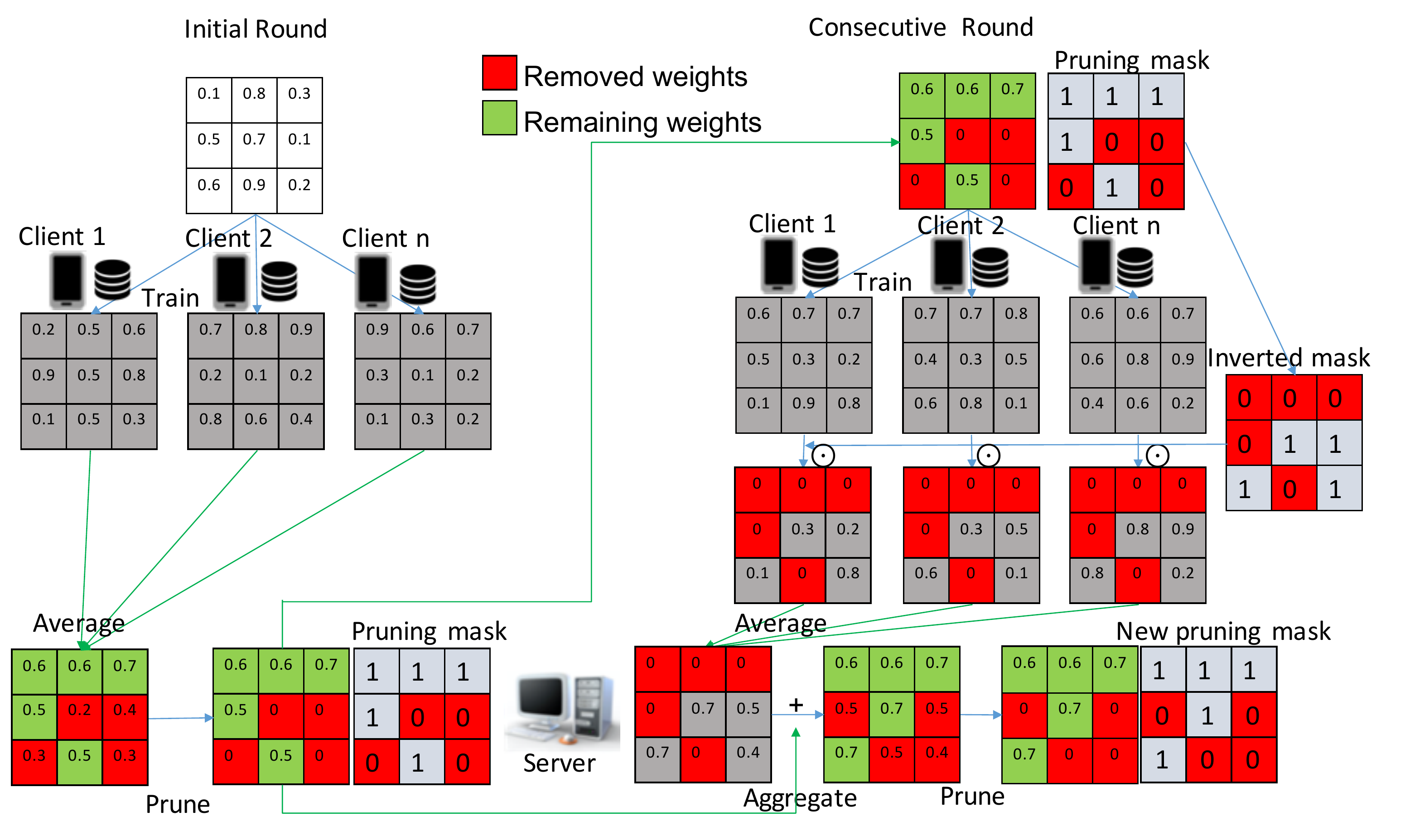} 
  \caption{Overview of Complement Sparsification in FL} 
  \label{overview}
\end{figure*}

The recent literature contains several works on model pruning for FL. An online learning approach~\cite{han2020adaptive} determines the near-optimal communication and computation trade-off by gradient sparsity. Liu et al.~\cite{liu2021adaptive} apply model pruning and maximize the convergence rate. PruneFL~\cite{jiang2022model} adapts the model size to find the optimal set of model parameters that learns the ``fastest''.  FL-PQSU~\cite{xu2021accelerating} is composed of a 3-stage pipeline: structured pruning, weight quantization, and selective updating. Yu et al.~\cite{yu2021adaptive} present an adaptive pruning scheme, which applies dataset-aware dynamic pruning for inference acceleration. In SubFedAvg~\cite{vahidian2021personalized}, the clients use a small subnetwork through pruning. Although most of these works achieve comparable model accuracy with vanilla FL and save some communication when the clients send the local updates to the server, they all impose substantial computation overhead on clients for additional optimizations or recovering the performance loss from pruning. In CS, pruning has very low computation overhead at the clients, as their only task is to remove the weights that were previously non-zero in the global sparse model. This low overhead makes CS practical for resource-constrained devices. 

While the works mentioned so far prune model weights, other works choose to remove neurons from the model at the clients. To cope with device heterogeneity, Ordered Dropout (OD)~\cite{horvath2021fjord} lets the clients train subnetworks of the original network in an ordered fashion. However, OD cannot save any communication from the server to the clients. In FedDrop~\cite{wen2022federated}, subnets are randomly generated from the global model at the server using dropout with heterogeneous dropout rates, and the clients only train and transmit the subnets to the server. This work saves communication bidirectionally, but suffers from inferior model accuracy compared to vanilla FL. CS not only reduces the bidirectional communication overhead, but also achieves comparable performance with vanilla FL.

In addition to pruning, there are other methods targeting the overhead in FL. The works in~\cite{wang2019adaptive, karimireddy2020scaffold} optimize the communication frequency. 
LotteryFL~\cite{li2020lotteryfl} communicates the personalized lottery networks learnt by the clients. 
Ozkara et al.~\cite{ozkara2021quped} use quantization and distillation for personalized compression in FL by manipulating the loss function at the clients. In~\cite{lin2018dgc}, the clients only send large gradients for aggregation and leave small gradients to accumulate locally until they become large enough. These works cannot enjoy all the benefits of using a sparse model, such as better generalization~\cite{bartoldson2020generalization} for a model to maintain good performance on unseen data, 
and higher robustness to adversarial attacks~\cite{10.1145/3386263.3407651,wu2020mitigating}. Since these methods belong to different classes of model compression, we do not compare them with CS.

%% file: methodology.tex
\section{Complement Sparsification in FL}
\label{sec:methodology} 

Complement Sparsification (CS) aims to reduce the bidirectional communication overhead between the server and the clients, impose minimum computation overhead on the system, and achieve good model performance. 
Figure~\ref{overview} shows its overview. In the initial round, the clients train from random weights and send their dense models to the server. After aggregation, the server prunes a percentage of model weights with low magnitude and sends the global sparse model to the clients. A pruning mask is also sent to the clients to mark the pruned weights. A 0 in the mask means the weight is removed, while a 1 means the weight remains. In the following rounds, after training, the clients apply the inverted mask of the global sparse model and send their sparse models back. The server aggregates the client models with the global sparse model from the previous round. Because the inverted mask keeps the weights of the client models that were originally zero in the global sparse model, a full dense model is produced by the aggregation. In the new dense model, the weights with low magnitude are pruned away, and a new global sparse model is produced with a new pruning mask different from the one in the previous round. 
The new model has a different subset of non-zero weights because the client model weights are amplified with a given aggregation ratio to outgrow other weights. 

The accuracy of the model improves over time, as all the model weights get eventually updated. Unlike pruning methods that require fine-tuning, the computation overhead of CS is merely removing some weights. The bidirectional communication overhead is also substantially reduced because both the server and the clients transfer sparse models. 

\subsection{Preliminaries}

In order to formulate CS, we start with the formulation of FL, which is a distributed DL system that finds the model weights $w$ that minimize the global empirical loss $F(w)$:
\begin{equation}
\label{eq:opt}
\min_{w}{F(w)} : = \sum_{n=1}^{N} \frac{|x_n|}{|x|} F_n(w) 
\end{equation} 

\begin{equation}
\label{eq:local_risk}
F_n(w) := \frac{1}{|x_n|}\sum_{i\in x_n} f_i(w)
\end{equation} 

where $F_n(w)$ is the local empirical loss for each client $n\in \{1,2,..,N\}$, $x_n$ is the local dataset of client $n$, $|x_n|$ is the dataset size of client $n$, $|x| = \sum_{n=1}^{N} |x_n|$ is the dataset size of all clients, and $f_i(w)$ is the loss function of a given client for a given data sample $i$ in its dataset.

Each client $n$ trains on its local data in every round.
\begin{equation}
  \label{eq:local_train_va}
\theta_{t+1,n} = w_{t} -  \eta g_n
\end{equation}

where $\theta$ is the current local model, $w_t$ is the global model of previous round, $\eta$ is the learning rate, and $g_n = \nabla F_n(w_t) $ is the average gradient of $w_t$ on its local data. This step may iterate multiple times with different batches of data, and repeat over the whole dataset.

Without loss of generality, we assume that every client participates in aggregation in every round. The server aggregates the learning outcomes from the clients as shown in equation either (\ref{eq:fl_va}) or (\ref{eq:fedavg}). 
\begin{equation}
  \label{eq:fl_va}
w_{t+1} = w_t  - \eta \sum_{n=1}^{N} \frac{|x_n|}{|x|} g_n
\end{equation}

\begin{equation}
  \label{eq:fedavg}
w_{t+1} = \sum_{n=1}^{N} \frac{|x_n|}{|x|} \theta_{t+1,n}
\end{equation}

Equations (\ref{eq:fl_va}) and (\ref{eq:fedavg}) are equivalent because of (\ref{eq:local_train_va}). In (\ref{eq:fl_va}), the server can use a different learning rate $\eta$ from the client learning rate $\eta$ in (\ref{eq:local_train_va}). 

\subsection{CS Workflow}
\subsubsection{Initial Round.}
CS starts from vanilla FL. The aggregated weights $w_{t+1}$ are pruned by the server, with a pruning function $(w_{t+1}', mask) = Prune(w_{t+1})$. The pruning function in CS removes the weights with low magnitude without any deliberate fine-tuning. We choose it because of its low overhead.
The pruned model $w_{t+1}'$ and the pruning mask $mask$ are sent to the clients for the following rounds.  The pruning $mask$ is a binary tensor indicating where $w_{t+1}'$ has weights set to 0.

\subsubsection{Consecutive Rounds.} 
\label{subsec:round} 
In a new round, each client $n$ receives the pruned model $w_{t}'$ from the server, trains it on the local data $x_n$ , and produces a new local model $\theta_{t+1,n}$: 

\begin{equation}
  \label{eq:local_train}
\theta_{t+1,n} = w_{t}' - \eta g_n
\end{equation}

Next, the clients compute the inverted bit-wise $\neg mask$ and apply the element-wise product $\odot$ between $\neg mask$ and $\theta_{t+1,n}$ (equation~(\ref{eq:flip_pruning})). If we want to save communication overhead and not send the $mask$ from the server to the clients, the clients can derive $\neg mask$ directly from $w_{t}'$, at the expense of a trivial computation overhead.

\begin{align}
\theta_{t+1,n}' &= \theta_{t+1,n} \odot \neg mask   \label{eq:flip_pruning} 
\end{align}

The server receives the complement-sparsified weights $\theta_{t+1,n}'$ from clients and aggregates them with $w_{t}'$ and an aggregation ratio $\eta'$, as shown in equation~(\ref{eq:new_aggregate}). 

\begin{align}
w_{t+1} 
&= w_t' + \eta' \sum_{n=1}^{N} \frac{|x_n|}{|x|} \theta_{{t+1},n}' \label{eq:new_aggregate}
\end{align}

Then, the server repeats the protocol from the previous rounds, and the CS workflow continues iteratively.

\subsection{Algorithmic Description}

Algorithm~\ref{alg_cs} shows the pseudo-code of CS.  CS executes as a multi-round, iterative FL cycle (line 4-14), involving local model updates done by the clients with batches of data (lines 16-20), complement sparsifying the local models (line 21), server aggregation (lines 9-12), and the global model pruning (line 13). To prune the global model, we remove the weights with low magnitude (lines 26-29) and generate a binary tensor masking the zero weights (lines 30-33).

\begin{algorithm}[h]
    \scriptsize
	\caption{Complement Sparsification Pseudo-code~\label{alg_cs}} 
	\begin{algorithmic}[1]
 	    \Procedure{ServerExecute:}{}
                \State require CS aggregation ratio $\eta'$ and server model sparsity $p\%$
	        \State initialize $t=0$, $w_0$ \textbf{randomly}, and tensor $mask$ to \textbf{zero}
            \While {!converged}
         \State // Update Done at Clients and Returned to Server
         \For {each client $n$} // In Parallel
                    \State $(\theta_n, |x_n|)$ = n.\Call{ClientUpdate}{$w_t,mask$}
        \EndFor                    
                    %
                     \State $|x|=\sum_n|x_n|$
	                    \If{$t==0$}
	                        \State $w_{t+1} \gets \sum_{n=1}^{N}\frac{|x_n|}{|x|}\theta_{n}$
	                    \Else
	                        \State $w_{t+1} \gets w_t+\eta'\sum_{n=1}^{N}\frac{|x_n|}{|x|}\theta_{n}$
	                    \EndIf
                      \State $(w_{t+1}, mask) \gets \Call{Prune}{w_{t+1},p}$
         \State $t$++
         \EndWhile
	    \EndProcedure

	    \item[]
     	    \Procedure{ClientUpdate}{$w, mask$}
          \State // Executed at Clients
                    \State require step size hyperparameter $\eta$
                    \State $x_n \gets$ local data divided into minibatches
                    \For {each batch $b \in x_n$}
                    \State $\theta_{n} = w - \eta \nabla F_n(w;b)$
                    \EndFor
                    \State $\theta_{n} \gets \theta_{n} \odot \neg mask$
                    \State // Results Returned to Server
                    \State \Return $(\theta_{n}, |x_n|)$
	    \EndProcedure

     	    \item[]
     	    \Procedure{Prune}{$w,p$}
          \State //  Executed at Server
                    \State $th \gets$ $p^{th}$ percentile in $w$
                    \For{each element $e \in w$}
                        \If{$e<th$}
                            \State $e \gets 0$
                            \EndIf
                    \EndFor
                \State $mask \gets w$
                \For{each element $e \in mask$}
                        \If{$e!=0$}
                            \State $e \gets 1$
                            \EndIf
                            \EndFor
                 \State \Return $(w,mask)$
	    \EndProcedure
	\end{algorithmic} 
\end{algorithm} 

\subsection{Technical Insights}
In FL, the clients produce models that fit the local data, while the server's aggregation averages out the noise in the client models and produces a global model that fits the global data. In other words, the clients and the server are in a complementary relationship. In every round, the clients perturb the global model to follow their local data distribution better, and the server conciliates the client models to capture the global data distribution. CS draws from these insights when it creates complementary sparse models at the server and the clients, respectively. In this way, it can reduce the computation and communication overhead, while achieving good model performance.

In CS, the server extracts a sparse model from the aggregated dense model. This sparse model preserves the global data distribution. Although the server does not fine-tune the sparse model, the clients perform implicit fine-tuning. They learn the local data distribution and create client sparse models that reflect shifts between the local and the global distribution. The updates are more easily reflected in the complement set of the global sparse model weights (i.e., the weights that were previously 0). Therefore, the clients complement-sparsify the models as in equation (\ref{eq:flip_pruning}), and only send the important model updates to the server with low communication overhead. This process also avoids overfitting the non-zero weights of the global sparse model by the clients' local data. The computation overhead is mostly imposed on the server, as the clients merely apply the inverted pruning mask.

Because we want all the weights to get updated over time for an accurate model, in every round, CS needs to produce a full dense model and generate a pruning mask different from the previous round. This is achieved by aggregating the complementary weights of the client models at round $t+1$ with the global model weights at round $t$ as in equation (\ref{eq:new_aggregate}). More specifically, the new aggregated model weights are calculated by adding the global sparse model weights and the weighted sum of the client weights. The server uses a constant aggregation ratio $\eta'>1$ to ensure that the pruned weights from the previous round outgrow the other weights, thus, will be less likely to be pruned in the current round. 
If some client updates are always small and are consequently removed by the server, the training can use a higher $\eta'$, but $\eta'$ shall not be higher than $1/\eta$ to avoid gradient explosion (see section~\ref{subsec:theo}).

\subsection{Algorithm Analysis}
\label{subsec:theo}

To show that in terms of performance CS is indeed an approximation of vanilla FL, we derive the aggregation function of vanilla FL (\ref{eq:fl_va}) from CS (\ref{eq:new_aggregate}), as follows. 

\begin{align}
w_{t+1} &= w_t' + \eta' \sum_{n=1}^{N} \frac{|x_n|}{|x|} \theta_{{t+1},n}'  \tag{\ref{eq:new_aggregate} revisited}\\
&\approx w_t' + \eta' \sum_{n=1}^{N} \frac{|x_n|}{|x|} (\theta_{t+1,n} -w_t') \label{eq:s1} \\
& =  w_t' - \eta'\eta \sum_{n=1}^{N} \frac{|x_n|}{|x|} \frac{w_t'-\theta_{{t+1},n}}{\eta} \label{eq:s2} \\
& = w_t'- \eta'\eta   \sum_{n=1}^{N} \frac{|x_n|}{ |x|} g_n  \label{eq:s3}  \\
& \approx w_t- \eta'\eta  \sum_{n=1}^{N} \frac{|x_n|}{|x|} g_n \label{eq:s4} 
\end{align}

Equation (\ref{eq:s1}) is from $\theta_{t+1,n}'\approx \theta_{t+1,n}-w_t'$. This is because the locally trained client model $\theta_{t+1,n}$ differs from the previous global sparse model $w_t'$ mostly on the zero weights of $w_t'$.
$\theta_{t+1,n} -w_t'$ sets the non-zero weights in $w_t'$ to 0, similar with $\theta_{t+1,n} \odot \neg mask$ in (\ref{eq:flip_pruning}). Equation (\ref{eq:s2}) is derived by taking $-\eta$ out of the sum. 
Equation (\ref{eq:s3}) is derived by using (\ref{eq:local_train}) in (\ref{eq:s2}).
The final result in equation (\ref{eq:s4}) is because the pruned weights $w_t'$ approximate the weights before pruning $w_t$, as they only differ in the small magnitude weights.
Comparing (\ref{eq:s4}) with (\ref{eq:fl_va}), the server applies $\eta'\eta$ as its learning rate. 
The aggregation ratio $\eta'$ is essentially the server-client learning rate ratio used to adjust the server learning rate over the client learning rate. In practice, because learning rate is typically chosen between 0 and 1, $\eta'$ shall be chosen between 1 and $1/\eta$ to ensure $\theta_{{t+1},n}'$ outgrows $w_t'$ without exploding $w_{t+1}$.

%% file: results.tex
\section{Evaluation} 
\label{sec:results}

The evaluation has six goals: (i) Compare the learning progress of CS and vanilla FL; (ii) Compare the learning progress of CS and FL pruning baselines; (iii) Investigate the effectiveness of low overhead pruning in CS; (iv) Quantify the communication savings in CS; (v) Quantify the computation savings in CS; (vi) Investigate the trade-off between model sparsity and model performance in CS. 

\subsection{Datasets}

CS is evaluated with two benchmark datasets in LEAF~\cite{https://doi.org/10.48550/arxiv.1812.01097}: Twitter and FEMNIST. Twitter consists of 1,600,498 tweets from 660,120 users. We select the users with at least 70 tweets, and this sub-dataset contains 46,000+ samples from 436 users. FEMNIST consists of 80,5263 images from 3,597 users. The images are 28 by 28 pixels and represent 62 different handwritten characters (10 digits, 26 lowercase, 26 uppercase). We choose these two datasets because they represent important types of data in DL, text and image, and also allow us to observe how CS behaves with different scales of user pools and datasets. 

In LEAF, we can choose IID or non-IID sampling scenarios. To evaluate CS under more realistic conditions, we choose non-IID for both datasets and make sure the underlying distribution of data for each user is consistent with the raw data. The training dataset is constructed with 80\% of data from each user, and the rest of the data are for testing. 

\subsection{Models}

We use a sentiment analysis (SA) model for the Twitter dataset, which classifies the emotions as positive, negative, or neutral. For example, with the inferred emotions of mobile users' text data, a smart keyboard may automatically generate emoji to enrich the text before sending. Our SA model first extracts a feature vector of size 768 from each tweet with pre-trained DistilBERT~\cite{sanh2019distilbert}. Then, it applies two dense layers with ReLU and Softmax activation, respectively, to classify the feature vector. The number of hidden states of the two dense layers are 32 and 3, respectively. 

We use a CNN-based image classification (IC) model for the FEMNIST dataset. This model uses three  convolutional layers and two dense layers to classify an image into one of the 62 characters. The three convolutional layers have 32, 64, and 64 channels, respectively, with 3 by 3 filters, stride of 1, and ReLU activateion. A max pooling follows the first and the third convolutional layers. Then, the flattened tensor is fed into two dense layers of 100 and 62 neurons, with ReLU and Softmax activation, respectively. 

The SA model has 24,707 trainable parameters, while the IC model has 164,506 trainable parameters. We choose these DL models also because we want to observe whether CS performs differently on different model sizes. 

\subsection{Experimental Settings}

\begin{table*}[t]
\fontsize{9}{9}\selectfont
\centering

\begin{tabular}{|l|l|l|l|l|l|l|}
\hline
Model & Optimizer & \shortstack{Weight\\initializer} & \shortstack{Client\\LR} & \shortstack{Aggregation\\ratio} & \shortstack{Batch\\size} & Epoch  \\ \hline
SA  & Adam      & he\_uniform            & 0.01      & 1.5       & 64        & 5        \\ \cline{1-7}        
IC  & Adam      & he\_uniform            & 0.01      & 1.5       & 64        & 5    \\ \hline
\end{tabular}
\caption{Training hyper-parameters for SA and IC models} 
\label{table_model_settings}
\end{table*}

We implement CS with Flower~\cite{https://doi.org/10.48550/arxiv.2007.14390} and Tensorflow. The experiments are conducted on a Ubuntu Linux cluster (Intel(R) Xeon(R) CPU E5-2680 v4 @ 2.40GHz with 512GB memory, 4 NVIDIA P100-SXM2 GPUs with 64GB total memory). We tested CS with different hyper-parameters, and only present the convergence progress with the hyper-parameters that led to the best results.
Table~\ref{table_model_settings} shows the training hyper-parameters for the two models. 
We set the aggregation ratio ($\eta'$ in equation~\ref{eq:new_aggregate}) to 1.5 to avoid clients' training outcomes being pruned away if they are too small. 
We set the server model sparsity to 50\%, unless otherwise specified.

\subsection{Baselines}
We compare CS with two recently published baselines: PQSU~\cite{xu2021accelerating}, and PruneFL~\cite{jiang2022model}. PQSU is composed of structured pruning, weight quantization, and selective updating. PruneFL includes initial pruning at a selected client, further pruning during FL, and adapts the model size to minimize the estimated training time. 
As CS, both PQSU and PruneFL aim to reduce communication and computation overhead in FL, and assume the server has no access to any raw data. 

We run PruneFL from its GitHub repository. Since PQSU's orginal source code can not run continuous FL, we implement PQSU with Flower and Tensorflow, similar to CS. To make them comparable, we use the same data, model structures, model sparsity, and hyper-parameters. 

\subsection{Results}

\begin{figure}[t!]
\centering
\begin{minipage}[b]{0.23\textwidth}
\resizebox{1\textwidth}{!}{%
\begin{tikzpicture}
\begin{axis}[
    xlabel={Round},
    ylabel near ticks,
    xmin=0, xmax=51,
    ymin=0.3, ymax=0.8,
    legend pos=south east,
    ymajorgrids=true,
    grid style=dashed,
]
\addplot[color=red,smooth,mark=None,] table [x=round, y=FL_all, col sep=comma, mark=none, smooth] {data/sent1.csv};
\addlegendentry{vanilla FL all users}
\addplot[color=blue,smooth,mark=None,] table [x=round, y=CS_all, col sep=comma, mark=none, smooth] {data/sent1.csv};
\addlegendentry{CS all users}
\end{axis}
\end{tikzpicture}}
\caption{Test set accuracy vs. communication rounds for SA trained with all users in every round}
\label{SA_all}
\end{minipage}%
   \hfill
  \begin{minipage}[b]{0.23\textwidth}
\resizebox{1\textwidth}{!}{%
\begin{tikzpicture}
\begin{axis}[
    xlabel={Round},
    ylabel near ticks,
    xmin=0, xmax=51,
    ymin=0.3, ymax=0.8,
    legend pos=south east,
     legend style={font=\LARGE},  
    ymajorgrids=true,
    grid style=dashed,
]
\addplot[color=red,smooth,mark=None,] table [x=round, y=FL_10, col sep=comma, mark=none, smooth] {data/sent1.csv};
\addlegendentry{vanilla FL 10 users}
\addplot[color=blue,smooth,mark=None,] table [x=round, y=CS_10, col sep=comma, mark=none, smooth] {data/sent1.csv};
\addlegendentry{CS 10 users}
\addplot[color=green,smooth,mark=None,] table [x=round, y=prunefl, col sep=comma, mark=none, smooth] {data/sent1.csv};
\addlegendentry{PruneFL 10 users}
\addplot[color=black,smooth,mark=None,] table [x=round, y=pqsu, col sep=comma, mark=none, smooth] {data/sent1.csv};
\addlegendentry{PQSU 10 users}
\end{axis}
\end{tikzpicture}}
\caption{Test set accuracy vs. communication rounds for SA trained with 10 random users in each round}
\label{SA_10}
\end{minipage}%
\end{figure}

\begin{figure}[t!]
\centering
\begin{minipage}[t]{0.23\textwidth}
\resizebox{1\textwidth}{!}{%
\begin{tikzpicture}
\begin{axis}[
    xlabel={Round},
    ylabel near ticks,
    xmin=0, xmax=500,
    ymin=0.0, ymax=0.8,
    legend columns=1,
 legend style={font=\Large},  
    legend pos= south east,
    ymajorgrids=true,
    grid style=dashed,
]
\addplot[color=red,smooth,mark=None,] table [x=round, y=FL_8, col sep=comma, mark=none, smooth] {data/fem.csv};
\addlegendentry{vanilla FL 10 users}
\addplot[color=blue,smooth,mark=None,] table [x=round, y=CS_0.5_8, col sep=comma, mark=none, smooth] {data/fem.csv};
\addlegendentry{CS 10 users}
\addplot[color=green,smooth,mark=None,] table [x=round, y=adaptive, col sep=comma, mark=none, smooth] {data/fem.csv};
\addlegendentry{PruneFL 10 users}
\addplot[color=black,smooth,mark=None,] table [x=round, y=pqsu_0.5, col sep=comma, mark=none, smooth] {data/fem.csv};
\addlegendentry{PQSU 10 users}
\end{axis}
\end{tikzpicture}}
\caption{Test set accuracy vs. communication rounds for IC trained with 10 random users in each round}
\label{IC_10}
\end{minipage}%
   \hfill
  \begin{minipage}[t]{0.23\textwidth}
\resizebox{1\textwidth}{!}{%
\begin{tikzpicture}
\begin{axis}[
    xlabel={Round},
    ylabel near ticks,
    xmin=151, xmax=351,
    ymin=0.65, ymax=0.76,
    legend pos=south east,
    ymajorgrids=true,
    grid style=dashed,
]
\addplot[color=red,smooth,mark=None,] table [x=round, y=FL_8, col sep=comma, mark=none, smooth] {data/fem.csv};
\addlegendentry{vanilla FL 10 users}
\addplot[color=blue,smooth,mark=None,] table [x=round, y=CS_0.5_8, col sep=comma, mark=none, smooth] {data/fem.csv};
\addlegendentry{CS 10 users}
\end{axis}
\end{tikzpicture}}
\caption{Zoom in of rounds 150-300 from Figure~\ref{IC_10}}
\label{IC_zoomin}
\end{minipage}%
\end{figure}

\textbf{Comparison with vanilla FL.} Figure~\ref{SA_all} shows the SA accuracy over training rounds when all users participate in every training round. In terms of best performance, the accuracy of CS is comparable with vanilla FL (73.3\% vs. 76.1\%). The less than 3\% difference is the cost of the significant overhead reduction in CS, which will be shown later in this section. In the initial rounds, there is a gap in accuracy due to pruning and the fact that clients did not have yet time to recover the performance loss through local training. However, as FL proceeds, CS allows clients to implicitly fine-tune the pruned model. The accuracy gap between CS and vanilla FL gradually decreases, until overfitting occurs for CS. Nevertheless, the system can use the best model for inference.

In FL on mobile and IoT devices, however, it is more realistic that only a small portion of users participate in each training round due to resource constraints on the devices. Figure~\ref{SA_10} shows the SA accuracy over communication rounds when 10 randomly selected users participate in each training round. In terms of best accuracy, CS (74.3\%) competes with vanilla FL (76.9\%). An advantage of CS is that its learning curve fluctuates significantly less than vanilla FL. This is because the effect of non-IID data is alleviated by runing in CS, while it is fully observed in vanilla FL.
This phenomenon is further confirmed by the IC model.

Figure~\ref{IC_10} shows the IC model accuracy over communication rounds when 10 randomly selected users participate in each round of training, which is a more realistic case than all users participating in every round. Overall, the learning curves between CS and vanilla FL are close, with best accuracy of 72.5\% and 76.3\%, respectively. FEMNIST is a much larger dataset with more users than Twitter, and IC is a more complex model, with more possible output classes than SA. Therefore, it takes IC more rounds (up to 500) to converge.  Let us note that the overfitting in Figure~\ref{SA_all} does not appear in Figure~\ref{IC_10}. This is because a larger model is less likely to be overfitted by smaller amounts of information (partial participation of clients every round). Figure~\ref{IC_zoomin} shows a zoomed in portion of the graph in Figure~\ref{IC_10}. The results demonstrate that vanilla FL fluctuates more abruptly than CS during the training. This is an important advantage for CS in practice. In real-world FL over mobile or IoT devices, the data gradually accumulate as FL proceeds, but the system can not wait hundreds of rounds for a final best model or does not have a fully representative test dataset to select the best model for users. In CS, it is safe to distribute the latest model to users, while the latest model for vanilla FL may suffer from inferior accuracy. 

\textbf{Comparison with baselines.} Figures~\ref{SA_10} and~\ref{IC_10} also show the model accuracy comparison between CS and the baselines. For SA, PruneFL and PQSU reach best model accuracy of 71.5\% and 73.4\%, which are 2.8\% and 0.9\% lower than CS, respectively. On the larger FEMNIST dataset, the results of IC show a clearer advantage for CS. For IC, PruneFL and PQSU reach best model accuracy of 55.5\% and 57.9\%, which are 17\% and 15\% lower than CS, respectively.   

The original PruneFL paper~\cite{xu2021accelerating} show comparable performance with vanilla FL on the FEMNIST dataset. However, the experiments used only the data of 193 users (out of 3597), the 193 users were further mixed and treated as 10 ``super-users", and all these users participated in every round of training. We believe our experiments represent a more realistic scenario because we use all users and do not mix multiple users into a ``super-user". For PQSU, the over-optimization on clients overfits the global model quickly. Thus, PQSU cannot benefit from additional data and training rounds. To conclude, the baselines suffer from poor performance in realistic conditions for large datasets. 

Next, we present a qualitative discussion to explain that the baselines have higher overhead. For communication overhead reduction, PruneFL uses an adaptive process in which the model not only shrinks, but also grows to reach the final targeted model sparsity. During the communications rounds with a grown model, PruneFL has higher communication overhead than CS. PQSU, on the other hand, can only save communication overhead in one direction, when clients transfer the sparse model to server. When PQUS transfers the model from the server to the clients, the communication overhead is higher than CS. Therefore, for the same targeted model sparsity, both PruneFL and PQSU have higher communication overhead than CS. For computation overhead, PruneFL imposes additional computation overhead including importance measure, importance aggregation, and model reconfiguration,  while PQSU requires clients to further fine-tune their sparse models after the training. Thus, they also suffer from higher computation overhead than CS, because CS only needs to remove weights from the dense model.

\textbf{Global sparse model vs. aggregated dense model.} Figures~\ref{SA_agg_prune} and~\ref{IC_agg_prune} show the comparison between the global sparse model and the aggregated dense model (i.e., the model before sparsification in each round) in CS for SA and IC models. Overall, the global sparse model exhibits a smooth learning curve and outperforms the aggregated model. This demonstrates the effectiveness of the low-overhead model pruning in CS, which reduces communication overhead and maintains good model performance by removing weights in low magnitude. In CS, the aggregated model not only captures the global distribution, but also gets polluted by the noisy distribution shift induced from the clients data. In each round, simply removing the weights with low magnitudes from the newly aggregated model can effectively eliminate the noisy distribution shift, and the global sparse model can steadily learn the global data distribution.   

\begin{figure}[t!]
\centering
\begin{minipage}[b]{0.23\textwidth}
\resizebox{1\textwidth}{!}{%
\begin{tikzpicture}
\begin{axis}[
    xlabel={Round},
    ylabel near ticks,
    xmin=0, xmax=50,
    ymin=0.3, ymax=0.8,
    legend pos=south east,
    ymajorgrids=true,
    grid style=dashed,
]
\addplot[color=red,smooth,mark=None,] table [x=round, y=CS_aggregated, col sep=comma, mark=none, smooth] {data/sent1.csv};
\addlegendentry{Aggregated model}
\addplot[color=blue,smooth,mark=None,] table [x=round, y=CS_10, col sep=comma, mark=none, smooth] {data/sent1.csv};
\addlegendentry{Global sparse  model}
\end{axis}
\end{tikzpicture}}
\caption{Global sparse model vs. aggregated dense model accuracy for SA with 10 random users every round}
\label{SA_agg_prune}
\end{minipage}%
   \hfill
  \begin{minipage}[b]{0.23\textwidth}
\resizebox{1\textwidth}{!}{%
\begin{tikzpicture}
\begin{axis}[
    xlabel={Round},
    ylabel near ticks,
    xmin=0, xmax=251,
    ymin=0.57, ymax=0.72,
    legend pos=south east,
    ymajorgrids=true,
    grid style=dashed,
]
\addplot[color=red,smooth,mark=None,] table [x=round, y=CS_aggregated, col sep=comma, mark=none, smooth] {data/fem.csv};
\addlegendentry{Aggregated model}
\addplot[color=blue,smooth,mark=None,] table [x=round, y=CS_0.5_8, col sep=comma, mark=none, smooth] {data/fem.csv};
\addlegendentry{Global sparse model}
\end{axis}
\end{tikzpicture}}
\caption{Global sparse model vs. aggregated dense model accuracy for IC with 10 random users every round}
\label{IC_agg_prune}
\end{minipage}%
\end{figure}

\textbf{Client model sparsity.} Sparsity is the percentage of zero weights in the model. A model with high sparsity can save both computation and communication cost in FL. In CS, the client model applies the inverted pruning mask, but in practice the client model sparsity is much higher than the complementary percentage of the server model sparsity. This is because when a client trains the global sparse model, only a portion of the zero weights in the global sparse model gets updated.  
Tables~\ref{SA_sparsity} and~\ref{IC_sparsity} show the client model sparsity of SA and IC averaged over the number of rounds until they converge, while varying the server model sparsity. Let us note that we do not include the mask in the communication overhead, due to its small size.
The server model sparsity indicates the communication cost saving from the server to the clients, while the client model sparsity represents the saving from the clients to the server. In general, the client model becomes sparser when the server model is denser. The results also show that the layers with more parameters benefit more from CS, as they are sparser than the small layers. The results demonstrate a substantial reduction in the communication overhead. For example, in Table~\ref{SA_sparsity}, when the reduction in the communication from the server to the clients is 80\% (i.e., server model sparsity), for SA, the reduction in the communication from the clients to server is 81.2\%. 
We observe similar results for IC (Table~\ref{IC_sparsity}).

\begin{table}[t!]

\centering

\resizebox{0.48\textwidth}{!}{
\begin{tabular}{|l|l|l|l|l|l|}
\hline
& \multirow{2}{*}{\shortstack{ Model\\layer }}  & \multicolumn{4}{c|}{Server model sparsity}  \\ \cline{3-6} 
           &  & 0.5 & 0.6   & 0.7 & 0.8        \\ \cline{1-6} 
\multirow{3}{*}{\shortstack{ Client \\ model \\ sparsity }}   
&Dense (768$\times$32)       & 0.933  & 0.885   & 0.841   & 0.812 \\ \cline{2-6}
&Output (32$\times$3 )       & 0.887 & 0.851  & 0.833  & 0.810    \\ \cline{2-6} 
&\textbf{Full model}       & \textbf{0.932}                     & \textbf{0.884}  & \textbf{0.841}  & \textbf{0.812}  \\ \hline
\end{tabular}
}
\caption{Client sparsity vs. server sparsity for SA}
\label{SA_sparsity}
\end{table} 

\begin{table}[t!]

\centering

\resizebox{0.48\textwidth}{!}{
\begin{tabular}{|l|l|l|l|l|l|}
\hline
& \multirow{2}{*}{\shortstack{ Model\\layer }}  & \multicolumn{4}{c|}{Server model sparsity}  \\ \cline{3-6} 
           &  & 0.5 & 0.6   & 0.7 & 0.8        \\ \cline{1-6} 
\multirow{6}{*}{\shortstack{ Client \\ model \\ sparsity }}   
&\shortstack{Conv2D\\ ($3\times3\times32$) }              & 0.569 & 0.528  & 0.587  & 0.788 \\ \cline{2-6}
&\shortstack{Conv2D\\ ($32\times3\times3\times64$)}       & 0.842 & 0.788  & 0.800  & 0.900    \\ \cline{2-6} 
&\shortstack{Conv2D \\($64\times3\times3\times64$)}    & 0.917 & 0.837  & 0.791  & 0.868    \\ \cline{2-6} 
&\shortstack{Dense \\($64\times16\times100$)}             & 0.920 & 0.863  & 0.853  & 0.909    \\ \cline{2-6} 
&Output ($100\times62$)                    & 0.756 & 0.722  & 0.721  & 0.698    \\ \cline{2-6} 
&\textbf{Full model}    & \textbf{0.904} & \textbf{0.843}  & \textbf{0.828}  & \textbf{0.891}  \\ \hline
\end{tabular}
}
\caption{Client sparsity vs. server sparsity for IC}
\label{IC_sparsity}
\end{table} 

\begin{table}[t!]

\centering

\resizebox{0.48\textwidth}{!}{
\begin{tabular}{|l|l|l|l|l|l|}
\hline
& \multirow{2}{*}{\shortstack{ Model layer/\\FLOPs }}  & \multicolumn{4}{c|}{Server model sparsity}  \\ \cline{3-6} 
           &  & 0.5 & 0.6   & 0.7 & 0.8        \\ \cline{1-6} 
\multirow{3}{*}{\shortstack{ FLOPs \\saved\\(\%) }}   
&Dense/147744                       & 31.1  & 36.1   & 41.3   & 47.0 \\ \cline{2-6}
&Output/585                         & 29.1 & 34.5  & 40.5 & 46.3    \\ \cline{2-6} 
&\textbf{Full model/148329}        & \textbf{31.1} & \textbf{36.1}  & \textbf{41.3}  & \textbf{47.0}  \\ \hline
\end{tabular}
}
\caption{CS training FLOPs saving vs. server sparsity for SA}
\label{SA_flops}
\end{table} 

\begin{table}[t!]

\centering

\resizebox{0.48\textwidth}{!}{
\begin{tabular}{|l|l|l|l|l|l|}
\hline
& \multirow{2}{*}{\shortstack{ Model layer/\\FLOPs }}  & \multicolumn{4}{c|}{Server model sparsity}  \\ \cline{3-6} 
           &  & 0.5 & 0.6   & 0.7 & 0.8        \\ \cline{1-6} 
\multirow{6}{*}{\shortstack{FLOPs \\saved\\(\%) }}   
&Conv2D/1168224           & 19.0 & 24.3  & 32.9  & 46.3 \\ \cline{2-6}
&Conv2D/13381824        & 28.1 & 32.9  & 40.0  & 50.0    \\ \cline{2-6} 
&Conv2D/17916096        & 30.6 & 34.6  & 39.7  & 48.9    \\ \cline{2-6} 
&Dense/614700           & 30.7 & 35.4  & 41.7  & 50.3    \\ \cline{2-6} 
&Output/37386           & 25.1 & 30.6  & 37.2  & 43.1    \\ \cline{2-6} 
&\textbf{\shortstack{Full model/33118230}}    & \textbf{29.1} & \textbf{33.6}  & \textbf{39.6}  & \textbf{49.3}  \\ \hline
\end{tabular}
}
\caption{CS training FLOPs saving vs. server sparsity for IC}
\label{IC_flops}
\end{table}

\textbf{Training FLOPs savings.} To evaluate the reduction in the computation overhead at the clients, we compute the training FLOPs savings based on the server and client model sparsity. We consider the number of multiply-accumulate (MAC) operations performed by each layer for both the forward and the backward pass during the training. In the forward pass, the clients perform FLOPs on the non-zero weights received from the server. In the backward pass, the MAC operations are counted for both the hidden state and the derivative. The hidden state MAC operations are fully counted as FLOPs. 
For the derivative, only the MAC operations on weights with non-zero values are counted as FLOPs. Here, the non-zero weights include both the non-zero set of weights received from the server and the zero weights that are updated to non-zero by the client. Let us note that the inverted pruning mask is applied after clients training, and therefore it does not help with FLOPs savings. 

Tables~\ref{SA_flops} and~\ref{IC_flops} show the CS training FLOPs savings for both SA and IC, as a percentage of the FLOPs needed by vanilla FL. 
The values displayed in the second column of the tables are the training FLOPs of a single sample in vanilla FL. We compare them with CS training FLOPs under different server model sparsity. 
We observe that CS can save up to 49.3\% training FLOPs, and the savings increase as the server model sparsity becomes higher. Similar with the communication savings, the layers with more parameters save a higher percentage of FLOPs.

\textbf{Server model sparsity vs. model accuracy.} Figures~\ref{SA_acc} and~\ref{IC_acc} show how the model accuracy varies with the server model sparsity for SA and IC. Since the server model sparsity is a parameter that can be set to different values for different models, it allows the system operators to achieve the desired trade-off between the model accuracy and the reduction in communication/computation overhead. In general, the model performs better when the server model sparsity is low. The results show that for SA, even a sparsity of 90\% can lead to good performance (an accuracy deterioration of merely 2\% compared to 50\% sparsity). However, for IC, the sparsity should be kept to at most 70\% to achieve acceptable performance.  

\begin{figure}[t!]
    \centering
    \begin{minipage}[b]{0.23\textwidth}
    \resizebox{1\textwidth}{!}{%
\begin{tikzpicture}
\centering   
\begin{axis}[
    ybar,
    ymin=0.5,
    enlarge x limits=0.2,
    bar width=0.4cm,
    height=6cm,
    width=8cm, 
    legend style={
        rotate=45,
        at={(1,1)},
        anchor=north east,
        legend columns=1,
        legend rows=2,
    },
    xlabel={Server Model Sparsity},
    tick label style={font=\LARGE},
    symbolic x coords={
    50,
    60,
    70,
    80,
    90
    },
      xtick=data,  
]
\addplot coordinates {(50,0.7434026) (60,0.7460977)(70,0.725772)(80,0.72891635)(90,0.72274005)};
\end{axis} 
\end{tikzpicture}}
  \caption{Accuracy as a function of server sparsity for SA}
  \label{SA_acc}
  \end{minipage}%
   \hfill
  \begin{minipage}[b]{0.23\textwidth}
  \resizebox{1\textwidth}{!}{%
  \begin{tikzpicture}   
\centering   
\begin{axis}[
    ybar,
    ymin = 0.5,
    enlarge x limits=0.2,
    bar width=0.4cm,
    height=6cm,
    width=8cm, 
    legend style={
        rotate=45,
        at={(1,1)},
        anchor=north east,
        legend columns=1,
        legend rows=2,
    },
    xlabel={Server Model Sparsity},
    tick label style={font=\LARGE},
    symbolic x coords={
    50,
    60,
    70,
    80
    },
      xtick=data,  
]
\addplot coordinates {(50,0.725356) (60,0.71116)(70,0.67022)(80,0.5914)};
\end{axis} 
\end{tikzpicture}}
  \caption{Accuracy as a function of server sparsity for IC}
  \label{IC_acc}
  \end{minipage}%
\end{figure}


%% file: conclusion.tex
\section{Conclusion}
\label{sec:conclusion} 
This paper proposed Complement Sparsification (CS), a practical model pruning for FL that can help the adoption of FL on resource-constrained devices. In CS, the server and the clients create and exchange sparse and complementary subsets of the dense model in order to reduce the overhead, while building a good accuracy model. CS performs an implicit fine-tuning of the pruned model through the collaboration between the clients and the server. The sparse models are produced with little computational effort. We demonstrate that CS is an approximation of vanilla FL. Experimentally, we evaluate CS with two popular benchmark datasets for both text and image applications. CS achieves up to 93.2\% communication reduction and 49.3\% computation reduction with comparable performance with vanilla FL. CS also performs better than baseline models in terms of model accuracy and overhead.

\section*{Acknowledgements}
This research was supported by the National Science Foundation (NSF) under Grant No. DGE 2043104. Any opinions, findings, and conclusions or recommendations expressed in this material are those of the authors and do not necessarily reflect the views of NSF.